\documentclass[10pt,twocolumn,letterpaper]{article}

\usepackage[article]{iccv}      
\usepackage[accsupp]{axessibility}  
\usepackage{colortbl}
\usepackage{tikz}
\definecolor{tabfirst}{rgb}{1, 0.7, 0.7} 
\definecolor{tabsecond}{rgb}{1, 0.85, 0.7} 
\definecolor{tabthird}{rgb}{1, 1, 0.7} 

\definecolor{iccvblue}{rgb}{0.21,0.49,0.74}
\usepackage[pagebackref,breaklinks,colorlinks,allcolors=iccvblue]{hyperref}

\def\paperID{279} 
\def\confName{ICCV}
\def\confYear{2025}

\title{LightSwitch: Multi-view Relighting with Material-guided Diffusion}

\author{
Yehonathan Litman
\and
Fernando De la Torre
\and
Shubham Tulsiani
\and
Carnegie Mellon University
\and
\url{https://yehonathanlitman.github.io/light_switch}
}

\begin{document}

\twocolumn[{%
    \renewcommand\twocolumn[1][]{#1}%
    \maketitle
    \begin{center}
        \centering
        \captionsetup{type=figure}\includegraphics[width=\textwidth]{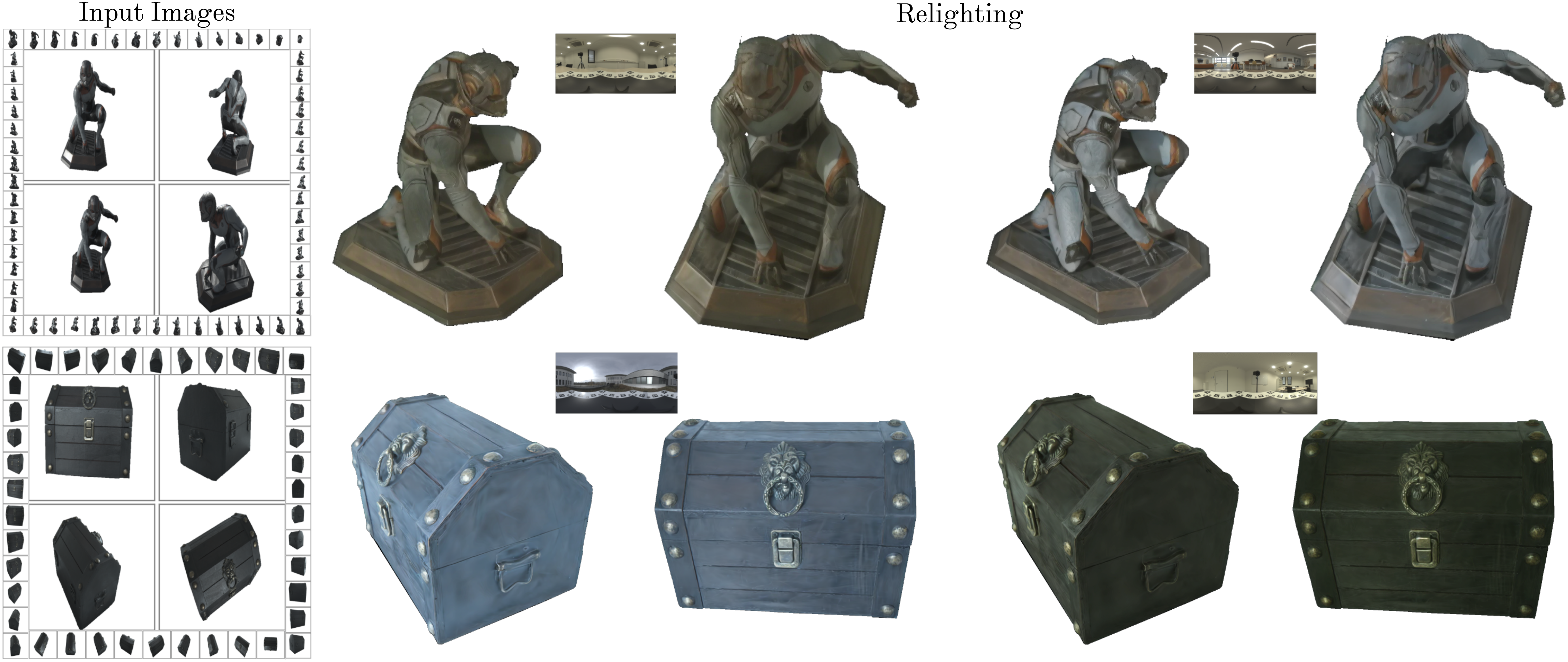}
        \captionof{figure}{
        \textbf{Consistent Multi-view via Material-Guided Relighting Diffusion.} We present LightSwitch, a framework for multi-view relighting. Given any number of input images under an unknown illumination, LightSwitch leverages multi-view attention and inferred material properties to predict consistent relighting, enabling applications for 2D and 3D relighting. 
        }
        \vspace{-0.1cm}

        \label{fig:teaser}
    \end{center}%
}]

\begin{abstract}
Recent approaches for 3D relighting have shown promise in integrating 2D image relighting generative priors to alter the appearance of a 3D representation while preserving the underlying structure. Nevertheless, generative priors used for 2D relighting that directly relight from an input image do not take advantage of intrinsic properties of the subject that can be inferred or cannot consider multi-view data at scale, leading to subpar relighting. In this paper, we propose Lightswitch, a novel finetuned material-relighting diffusion framework that efficiently relights an arbitrary number of input images to a target lighting condition while incorporating cues from inferred intrinsic properties. By using multi-view and material information cues together with a scalable denoising scheme, our method consistently and efficiently relights dense multi-view data of objects with diverse material compositions. We show that our 2D relighting prediction quality exceeds previous state-of-the-art relighting priors that directly relight from images. We further demonstrate that LightSwitch matches or outperforms state-of-the-art diffusion inverse rendering methods in relighting synthetic and real objects in as little as 2 minutes. 
\end{abstract}
\vspace{-0.25cm}

\section{Introduction}
\label{sec:intro}

We have witnessed impressive advances in the task of recovering 3D representations from multi-view captures, with methods like NeRF~\cite{mildenhall2020nerf} and Gaussian Splatting~\cite{kerbl20233d} allowing one to reconstruct generic objects or scenes easily. While these representations excel at modeling detailed geometry and appearance, they only seek to model the appearance within the capture environment, thus baking in lighting effects into the obtained representation. This prevents reconstructions from being imported and relit in novel environments for applications such as virtual reality or synthesizing visual effects. In this work, we aim to enable such relightable rendering of 3D representations under generic illumination and present an approach that given posed multi-view images, enables synthesizing novel relit views.

Current methods that tackle this relighting task can be categorized as either leveraging inverse rendering or learned relighting. In particular, the former class of methods~\cite{hasselgren2022nvdiffrecmc, litman2024materialfusion, liang2023gsir, zhang2022invrender, Jin2023TensoIR, zhenyuan2024bigs, R3DG2023} seek to infer a 3D representation disentangling geometry, appearance, and material properties, allowing relighting via physics-based rendering. However, these optimization-based approaches tend to be slow and their usage of (simple) differentiable renderers can limit their ability to model complex lighting effects. In contrast, `direct relighting' methods learn to directly generate a relit image a given (captured/rendered) source image and a target illumination. By adapting image diffusion priors, these methods can efficiently synthesize photo-realistic quality output in a feed-forward manner. However, these approaches operate on a single view, leading to inconsistencies in relighting across viewpoints. 

Our work also adopts a learning-based approach for direct relighting, with the key insight that instead of formulating this as a single-view relighting task, we should formulate it as one of consistently relighting \emph{multiple} input views. This can make the predictions consistent across views (making such a system better suited for 3D relighting), while also improving relighting performance as the cues observed in one view (\eg sharpness of a specularity) can inform the relighting in another. In addition to incorporating multi-view cues, we also draw inspiration from inverse rendering methods that benefit from understanding material properties and seek to leverage (predicted) material properties (intrinsics and albedo) as additional input.

We build on these insights and propose `LightSwitch', a relighting diffusion framework that produces multi-view consistent relighting informed by inferred intrinsic properties.  We validate LightSwitch on both synthetic and real-world data and find that leveraging multi-view and predict material cues yield significantly improved relighting compared to prior learning-based relighting methods. We also tackle the `3D relighting' task requiring synthesizing and consistently relighting a large set of query views, and design a distributed inference scheme for efficient inference. We show that when compared to state-of-the-art inverse rendering methods, LightSwitch allows improved/comparable relighting, while being significantly faster.
\section{Related Work}
\label{sec:related_work}

\paragraph{Image-based 3D Reconstruction.}

We have witnessed impressive recent progress in the task of recovering 3D from images. In particular representations like NeRF \cite{mildenhall2020nerf}, Gaussian Splatting \cite{kerbl20233d}, and their variants \cite{barron2023zipnerf, barron2022mipnerf360, verbin2024nerf, mueller2022instant, Huang2DGS2024, lyu20243dgsr, yan2024multi} allow representing detailed geometry and appearance, and can be inferred from multi-view images of generic objects and scenes. More recent approaches, often leveraging generative priors, even allow inferring such detailed outputs from sparse or single-view input \cite{zhou2023sparsefusion, Liu_2024_CVPR, voleti2024sv3d, wu2023reconfusion, long2023wonder3d, chen2024mvsplat360}, allowing end users to easily capture 3D. While these obtained reconstructs can capture the rich details of underlying 3D scenes, they are only capable of modeling the static environment observed in the images and cannot be imported to novel environments where their appearance would change under a different environmental illumination.

\paragraph{Relighting via Inverse Rendering.}

To enable relighting under novel environments, some works utilize inverse rendering to recover a 3D relightable asset. These approaches model intrinsic properties \cite{Munkberg_2022_CVPR, zhang2022invrender, liang2023gsir, hasselgren2022nvdiffrecmc, zhang2021nerfactor, Jin2023TensoIR, R3DG2023, litman2024materialfusion} or light transport effects ~\cite{nerv2021, zhenyuan2024bigs, boss2021neuralpil} to infer a factorized 3D representation that explains the observed images, allowing for relighting under novel illuminations. However, recovering intrinsics from a set of multi-view images is a non-trivial task, given that many different combinations of the intrinsic properties can be composed together to get the appearance in the source images. Some approaches use data-driven prediction of intrinsics \cite{chen2024intrinsicanything, kocsis2023intrinsic} to aid this, for example MaterialFusion \cite{litman2024materialfusion}, whose material model we adapt that uses a material prior model to aid with relighting. Nevertheless, inverse rendering pipelines are forced to rely on simple material models together with simpler lightweight differentiable renderers due to the computational constraints of physically-based renderers. Even so, optimization is still slow as real-time rendering requires considerable computation, making inverse rendering approaches time-consuming.

\begin{figure*}
    \centering
    \includegraphics[width=\textwidth]{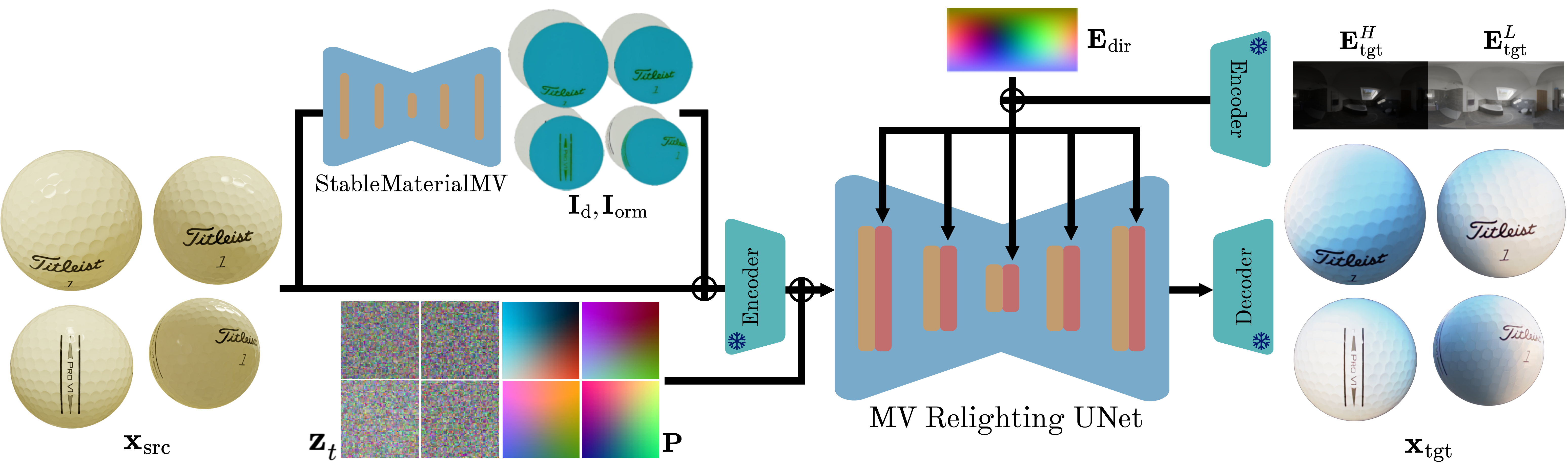}
    \caption{\textbf{LightSwitch Material-Relighting Diffusion Framework.} LightSwitch relights multi-view posed input images to a given target illumination. It infers and encodes multi-view consistent material image maps ($\mathbf{I}_\text{d}, \mathbf{I}_\text{orm}$) using a material diffusion model (StableMaterialMV \cite{litman2024materialfusion}) and concatenates them to the Pl\"{u}cker ray maps ($\mathbf{P}$), encoded input images ($\mathbf{x}_\text{src}$), and noisy latents ($\mathbf{z}_t$) in the channel dimension. The multi-view relighting UNet denoises the noisy latents and cross-attends to the lighting latents concatenated with the latent lighting directions ($\mathbf{E}_\text{dir}$). The lighting latents are encoded from the processed target environment map images $(\mathbf{E}^H_\text{tgt}, \mathbf{E}^L_\text{tgt})$. The Stable Diffusion encoders and decoder are kept frozen.}
    \label{fig:2d_approach}
\end{figure*}

\paragraph{Learning Direct Relighting.}

An alternate relighting approach that allows for high fidelity relighting is to learn a model that directly predicts relit images, in particular by adapting generative diffusion priors for high quality generation. These direct feed-forward approaches allow for predicting accurate relighting in little time for high resolution images while generalizing to unseen instances \cite{jin2024neural_gaffer, zhao2024illuminerf, zeng2024dilightnet}. However, not incorporating cues about the underlying material composition means the model is not using information that can help accurately relight assets with complex appearance effects. Even so, direct relighting models only take single-view images, limiting their practicality for 3D relighting where multi-view data provides important cues on the asset's inherent properties.

In contrast to aforementioned works, our proposed relighting diffusion framework incorporates intrinsic and multi-view cues to efficiently produce a high quality relighting directly from input images. The addition of these components to the relighting diffusion model significantly improves relighting for 2D and 3D scenarios. While some concurrent works have examined the task of recovering multi-view consistent relighting, they either do not leverage the prior knowledge of diffusion models \cite{zhang2024relitlrm}, rely on multi-illuminated data \cite{alzayer2024generativemvr}, or incorporate material information cues for video relighting only \cite{DiffusionRenderer, fang2025relightvid}.
\section{Methodology}
\label{sec:methodology}

\begin{figure*}
    \centering
    \includegraphics[width=\textwidth]{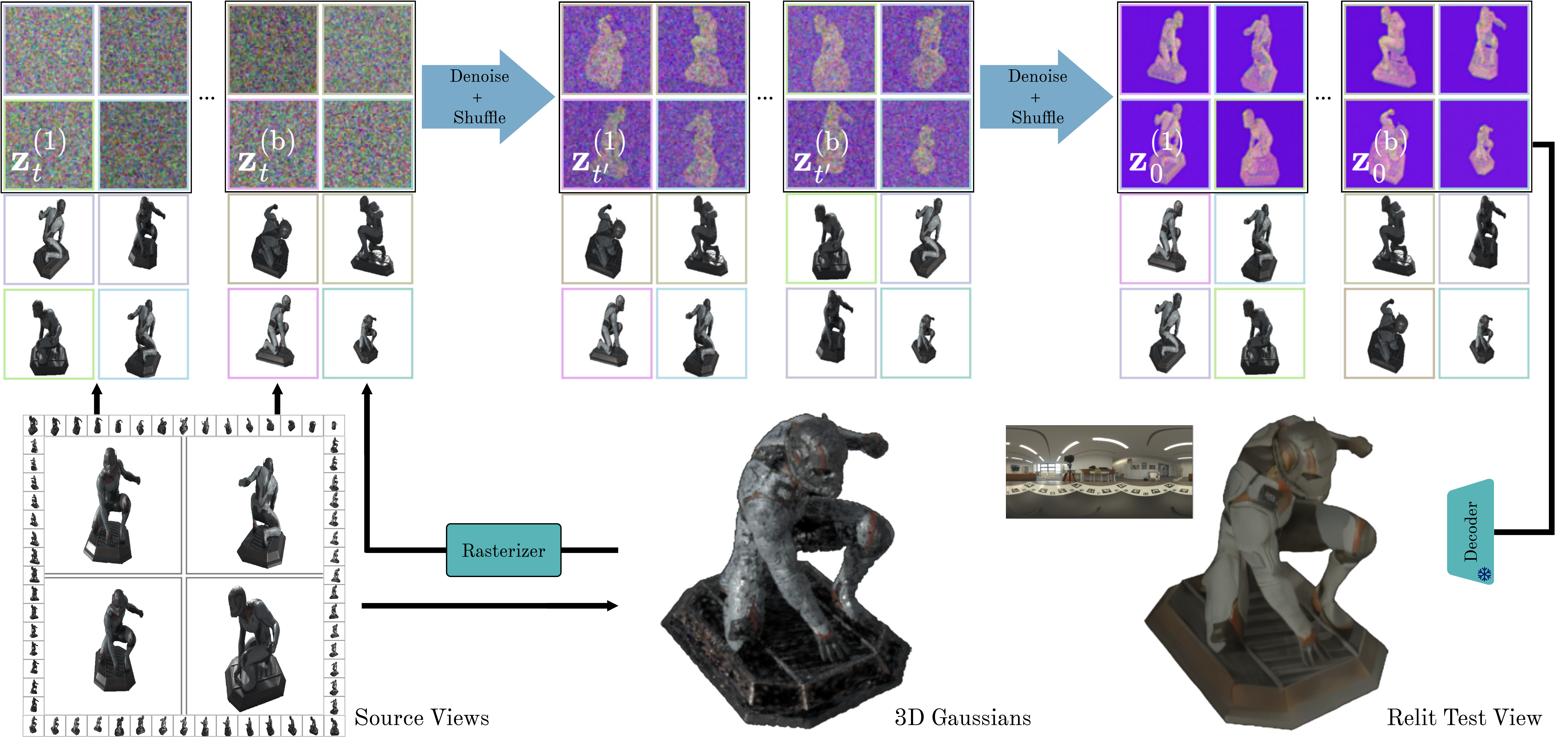}
    \caption{
    \textbf{Denoising Relighting for Any Number of Views.} Given the quadratic complexity of all-pair multi-view attention, we divide the input latents $\mathbf{z}_t$ into mini-batches $\mathbf{z}_t^\text{(1)}, \dots,  \mathbf{z}_t^\text{(b)}$ and make latents attend to each other only within a subset per denoising iteration. When the batches are shuffled after the denoising step, they can attend to another subset in the next iteration. By continuously shuffling the subsets across DDPM iterations we approximate the full relighting diffusion. To relight a novel view, we optimize a 3D gaussian splat on the training images and render the novel view using the rasterizer. The novel view is then inserted into the set source views, enabling consistent novel view relighting.
    }
    \label{fig:3d_approach}
\end{figure*}

In this section we introduce our diffusion framework, shown in Fig.~\ref{fig:2d_approach}, that takes in multi-view posed input images captured under fixed unknown illumination and relights them to a given target illumination. We adapt a text-conditioned diffusion model and finetune it for multi-view material-guided relighting and describe this in Sec. \ref{sec:2d_relighting}. We then introduce an efficient relighting denoising scheme at inference for 3D relighting in Sec. \ref{sec:3d_relighting}, shown in Fig.~\ref{fig:3d_approach}.

\subsection{Relighting Architecture}
\label{sec:2d_relighting}

Our goal is to design an architecture that can consistently relight a set of input images given a target environment map. Towards this, we seek to adapt the deep image priors contained in large scale diffusion models like Stable Diffusion 2.1 \cite{rombach2021highresolution}. Unlike previous approaches which focus on single-view relighting, we propose a model that allows multi-view consistent relighting. Moreover, in addition to conditioning on target illumination information, we also rely on material information cues for better relighting performance. We train this material-guided relighting diffusion model in stages beginning from single-view to yield a relighting model that synthesizes high quality multi-view consistent relightings.  

\paragraph{Material Aware Single-view Relighting.}

The relighting diffusion UNet, initialized from Stable Diffusion 2.1's UNet, is first finetuned to relight single RGB input views captured under unknown lighting $\mathbf{x}_\text{src}$ to target images $\mathbf{x}_\text{tgt}$ illuminated by a given target lighting $\mathbf{E}_\text{tgt}$. We modify the UNet's input layer to condition it on input images, material intrinsics, and camera pose information encoded as Pl\"{u}cker coordinate ray maps $\mathbf{P}$. To incorporate intrinsic material cues, the model utilizes per-pixel image material maps $\mathbf{I}_\text{d}, \mathbf{I}_\text{orm}$ corresponding to the albedo, occlusion, roughness, and metallicness (ORM)\footnote{The occlusion map is not used and set to zero.} components of the rendered image. The material representation used by the relighting model follows a simplified Disney principled BRDF model \cite{burley2012physically}, where each pixel in $\mathbf{I}_\text{d}, \mathbf{I}_\text{orm}$ contain an albedo $\mathbf{a} \in \mathbb{R}^{H \times W\times 3}$, roughness $r \in \mathbb{R}^{H \times W\times 1}$, and metallicness $m \in \mathbb{R}^{H \times W\times 1}$. The underlying material information is constant across illuminations and using it as conditioning lets the relighting model understand how to relight views with diverse appearance effects such as specularities and absorption. 

\paragraph{Incorporating Lighting.}

To incorporate lighting information into the denoising UNet, we add a lighting cross-attention module that attends to illumination information and finetune it together with the rest of the UNet. The target illumination information $\mathbf{E}_\text{tgt}$ is given initially as a high dynamic range image and transformed to two environment maps following \cite{jin2024neural_gaffer}: $\mathbf{E}^H_\text{tgt}$, which is the normalized environment map, and $\mathbf{E}^L_\text{tgt}$ which is tonemapped. The combination of these two maps helps inform the network about strong (e.g. lights, sun, etc.) and softer lighting (e.g. reflections, ambient lighting, etc.). These images are encoded using the Stable Diffusion encoder $\mathcal{E}$ and a directional embedding map $\mathbf{E}_\text{dir}$ is also produced corresponding to the per latent lighting direction. The lighting module attends to the concatenation $(\mathcal{E}(\mathbf{E}^H_\text{tgt}), \mathcal{E}(\mathbf{E}^L_\text{tgt}), \mathbf{E}_\text{dir})$.

At each training iteration we sample $(\mathbf{x}_\text{src}, \mathbf{x}_\text{tgt}, \mathbf{I}_\text{d}, \mathbf{I}_\text{orm}, \mathbf{P}, \mathbf{E}_\text{tgt})$ and finetune the diffusion model to denoise the noisy target view latent code $\mathbf{z}_t$. $\mathbf{z}_t$ is obtained by sampling a diffusion timestep $t$ and a 4-channel random noise $\epsilon$ and adding the noise to $\mathcal{E}(\mathbf{x}_\text{tgt})$. We concatenate $(\mathcal{E}(\mathbf{x}_\text{src}), \mathcal{E}(\mathbf{I}_\text{d}), \mathcal{E}(\mathbf{I}_\text{orm}), \mathbf{P})$, denoted as $\mathbf{C}$, along the channel dimension of $\mathbf{z}_t$ as conditioning and process $\mathbf{E}_\text{tgt}$ for cross attention. The diffusion loss for the single-view training stage is
 
\begin{equation}
    \mathcal{L}_{\text{diff}} = 
    \| \epsilon_\theta\left(\mathbf{z}_t, t ; \mathbf{C} \right) - \mathbf{v}_t \|_2^2,
\end{equation}

where $\mathbf{v}_t$ corresponds to v-prediction loss \cite{salimans2021progressive}. 

\paragraph{Denoising Multi-view Relighting.}

After finetuning our relighting diffusion model to relight single-view posed images to a target illumination given groundtruth material information, we continue finetuning it for multi-view prediction. We modify the denoising UNet with multi-view self-attention modules and continue training. By first training the model for single-view relighting, it forms an initial understanding of lighting interaction between the environment and object that is considerably harder to model in multi-view. Previous work~\cite{wang2023imagedream, shi2023mvdream} has shown that multi-view attention can be easily implemented in the self-attention module to enable consistent prediction across multiple views. Given a batch containing $k$ image latents, the self-attention layer consolidates the batch together such that every latent pixel across the batch is in the same space and can attend to all other latent pixels. When integrating this into the relighting diffusion model, it predicts a consistent most probable relighting for a given illumination across all input appearance information. This is replicated for the material diffusion model for predicting the most probable material maps across given input views. We show that including the multi-view attention module significantly boosts relighting quality.

We sample $k$ source views per object and use the following diffusion loss

\begin{equation}
    \mathcal{L}^{\text{mv}}_{\text{diff}} = 
    \| \epsilon_\theta\left(\mathbf{z}_t^{1:k}, t ; \mathbf{C}^{1:k} \right) - \mathbf{v}_t \|_2^2,
\end{equation}
We enable classifier-free guidance by setting $\mathcal{E}(\mathbf{E}^H_\text{tgt}), \mathcal{E}(\mathbf{E}^L_\text{tgt})$ to all zeros with a 10\% probability and set the guidance scale to 3. $k$ is set to 4 during training. Once the base multi-view model is trained, we continue training an upscaled model at a resolution of 512$\times$512 to produce higher fidelity relighting. 

\subsection{Lightswitch -- Scalable Efficient 3D Relighting}
\label{sec:3d_relighting}

With the trained upscaled multi-view LightSwitch diffusion model, we wish to apply it to 3D novel view relighting given sparse or dense multi-view data. In novel view relighting, training views of an asset are given as input and we wish to render an unseen view under a desired novel illumination. This can be practically challenging if the number of input views is too high as the compute requirement of transformers scales quadratically, and batch-wise processing can lead to inconsistencies. To address this, we introduce an efficient denoising mechanism that scales to an arbitrary number of input views as shown in Fig.~\ref{fig:3d_approach}. To enable novel relighting, we first render novel views under source illumination using novel view synthesis. We then input both the source views and synthesized novel views to our relighting network, allowing it to relight query views while using cues from the observed source views. 

\paragraph{Distributed Multi-view Relighting.}
At inference time, we can effectively denoise more images by shuffling the data and sampling new batches per denoising iteration. Over enough iterations, each latent attends to all other latents across the entire dataset, making the final prediction consistent throughout. Additionally, we can distribute the denoising step in parallel across compute to proportionally accelerate the diffusion process. This keeps the consistency of the final prediction and allows for fast scalable relighting on high resolution images, enabling highly accurate relighting as seen in the example output in Fig.~\ref{fig:teaser}. The shuffling and denoising procedure is applied to both the material and relighting diffusion models for efficient material and relighting inference. 

\paragraph{Rendering Test Views.}

Our approach distributes multi-view relighting effectively but can only relight given views. Thus, to relight novel views not included in the data, we optimize a 3D gaussian splat \cite{kerbl20233d} on the training data, render and encode the test view, and include it in the training data being denoised. Given the strong performance of novel view synthesis approaches we can sample a high quality test view and easily relight it with LightSwitch, enabling low latency novel view relighting. 

\begin{figure*}
    \centering
    \includegraphics[trim={0 0cm 0 0cm},clip,width=0.78\textwidth]{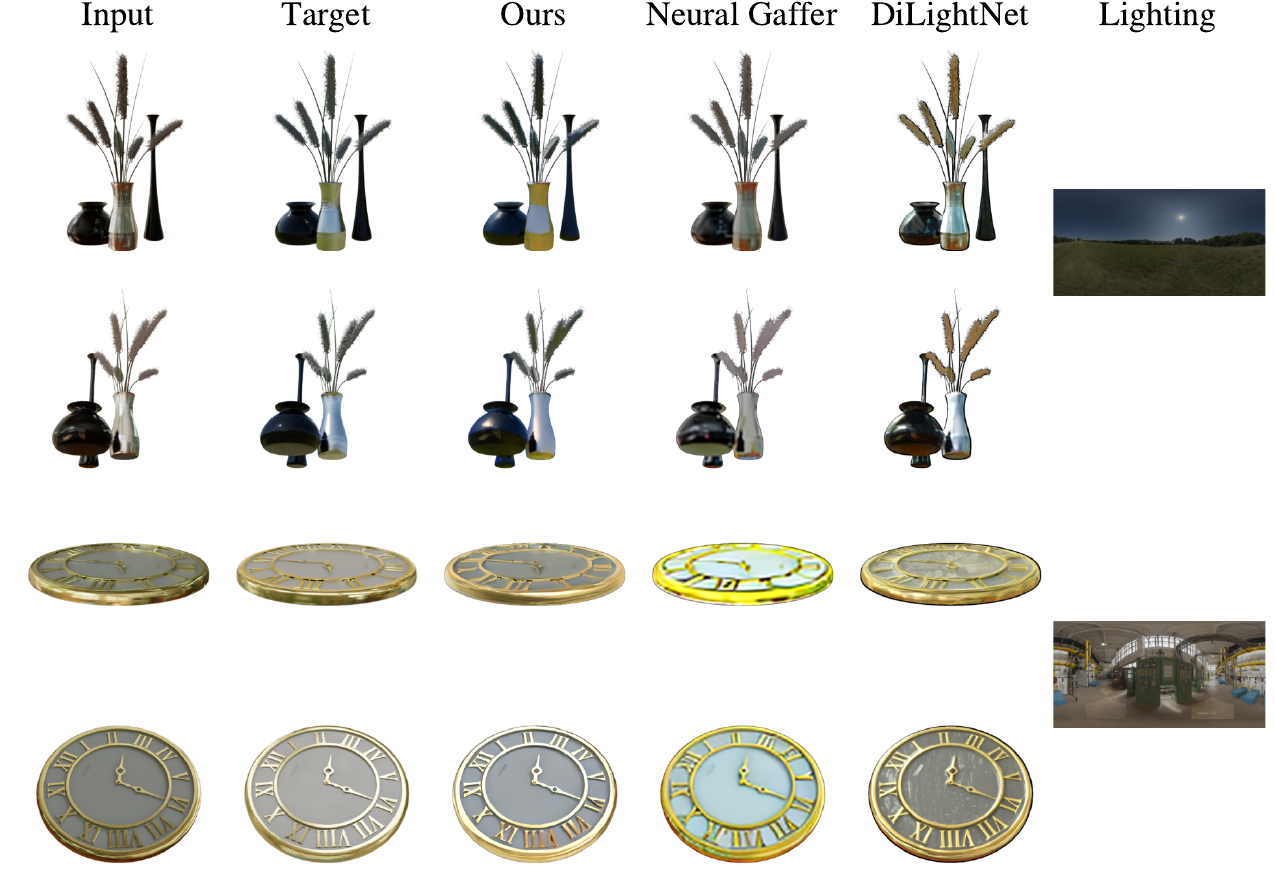}
    \caption{\textbf{Direct Relighting Comparison on Synthetic Objects.} Given 8 images of an object, LightSwitch predicts a multi-view consistent relighting under a target illumination. With its usage of inferred material information, our model accurately relights objects with complex appearance effects such as specularities. On the other hand, the baselines bake in details from the source view into the target relighting and relight inconsistently across views.
    }
    \label{fig:2d_relighting_comparison}
\end{figure*}
\label{sec:experiments}
\begin{table*}[!t]
    \centering
    \begin{tabular}{lccc|ccc|c}\toprule
     & \multicolumn{3}{c}{\bfseries Image Relighting (ILR)} & \multicolumn{3}{c}{\bfseries Image Relighting (SLR)} & {\bfseries Quality Drop} \\ \cmidrule(lr){2-4} \cmidrule(lr){5-7} \cmidrule{8-8}    \bfseries {Method} & PSNR $\uparrow$ & SSIM $\uparrow$ & LPIPS $\downarrow$ & PSNR $\uparrow$ & SSIM $\uparrow$ & LPIPS $\downarrow$ & PSNR $\downarrow$\\ \midrule
        DiLightNet \cite{zeng2024dilightnet} & \cellcolor{tabthird}23.84 & \cellcolor{tabthird}0.861 &  \cellcolor{tabsecond}0.238 & \cellcolor{tabthird}23.35 & \cellcolor{tabthird}0.859 & \cellcolor{tabsecond}0.238 & \cellcolor{tabthird}0.49\\
        Neural Gaffer \cite{jin2024neural_gaffer} & \cellcolor{tabsecond}24.34 & \cellcolor{tabsecond}0.883 & \cellcolor{tabthird}0.271 & \cellcolor{tabsecond}24.08 & \cellcolor{tabsecond}0.882 & \cellcolor{tabthird}0.272 & \cellcolor{tabsecond}0.26\\
        Ours & \cellcolor{tabfirst}26.01 & \cellcolor{tabfirst}0.888 & \cellcolor{tabfirst}0.216 & \cellcolor{tabfirst}25.86 & \cellcolor{tabfirst}0.885 & \cellcolor{tabfirst}0.215 & \cellcolor{tabfirst}0.15\\
        \midrule
        Ours (GT Materials) & 28.29 & 0.901 & 0.203 & 28.20 & 0.901 & 0.203 & 0.09\\
        \bottomrule
    \end{tabular}
    \caption{\textbf{Direct Relighting Accuracy Comparison.} We report the performance of our approach for 2D relighing on synthetic object data against other diffusion-based relighting methods. ILR corresponds to image level rescaling where the images are individually rescaled against the groundtruth image. SLR corresponds to scene level rescaling where we rescale using the average scale for all views per object. In each column, the \colorbox{tabfirst}{best}, \colorbox{tabsecond}{second best}, and \colorbox{tabthird}{third best} results are marked.}
    \label{tab:2d_relighting_comparison}
\end{table*}
\section{Experiments}

We evaluate LightSwitch on image sets of diverse assets captured under varied illumination to showcase our approach's performance in 2D and 3D scenarios. To show LightSwitch's relighting performance and effective denoising scheme in 2D, we conduct a comparison against other diffusion-based relighting priors on a held out synthetic object test dataset relit with unseen target illuminations. We then evaluate our 3D novel view relighting method on both synthetic and real objects to highlight its generalization, efficiency, and relighting capabilities. This is compared against inverse rendering methods that also enable novel view relighting. Lastly, we ablate our LightSwitch to showcase how the choice of integrating material and multi-view information aids relighting. 

\subsection{Experimental Setup}
\paragraph{Dataset.}

We curate a dataset of \raisebox{0.5ex}{\texttildelow}100K objects from a mixture of the BlenderVault \cite{litman2024materialfusion} and Objaverse~\cite{deitke2023objaverse, huang2024materialanything} data that was filtered to include high quality objects containing PBR maps. We sample 8 camera poses on a hemisphere around an object and render those views under 8 different environment maps. The environment maps are randomly selected from a dataset acquired from online sources such as Polyhaven and Laval ~\cite{laval_hdr}, giving \raisebox{0.5ex}{\texttildelow}4K environment maps that are also flipped and rotated randomly during training. At test time, we employ StableMaterialMV~\cite{litman2024materialfusion}, taken off the shelf, to infer materials from the input images to condition the relighting diffusion model. Given that StableMaterialMV was trained on 256$\times$256 images, we separately finetune it further to 512$\times$512 to estimate high quality material maps.

\paragraph{Metrics.} 

For both the 2D and 3D relighting evaluations, we account for the underlying albedo scale ambiguity by a scale against the groundtruth image before computing the PSNR, SSIM, and LPIPS metrics. We also report the approximate time from start to finish LightSwitch and other methods need to predict relightings. Similarly to previous work, the final results are computed as the mean across views for all objects.

\subsection{Image to Image Relighting}
\begin{figure*}
    \centering
    \includegraphics[width=\textwidth]{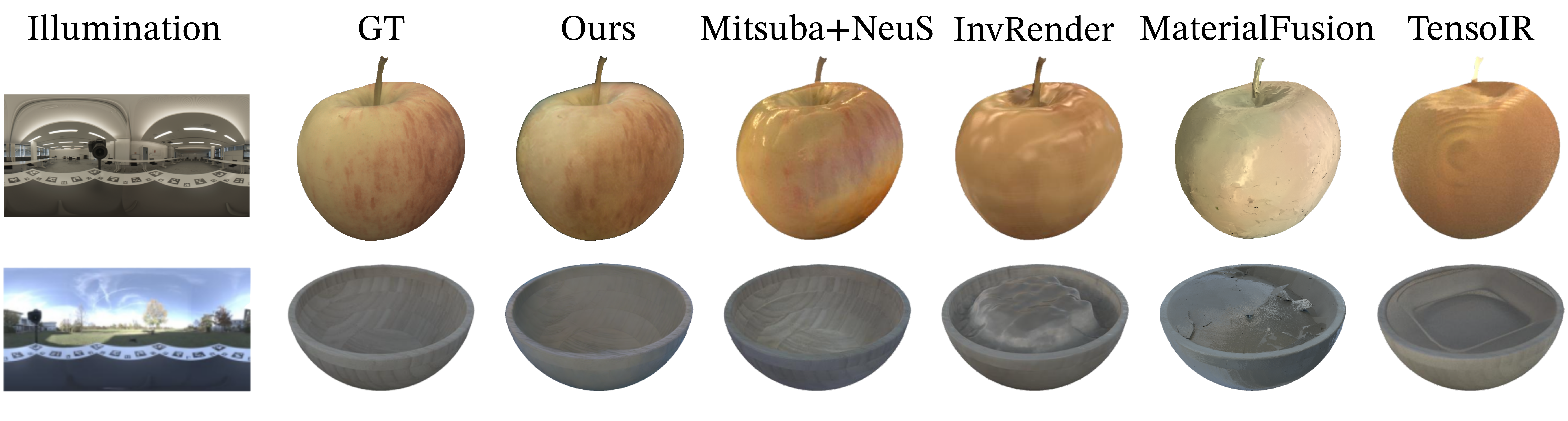}
\vspace*{-7mm}
    \caption{\textbf{3D Relighting Comparison on Objects With Lighting.} Our method successfully relights a novel view to a target illumination while the baselines exhibit errors in the relit appearance. LightSwitch's efficiency means it can relight a given novel view in 5 minutes at a high accuracy while operating on images at the original resolution (1728$\times$1120).
    }
    \label{fig:3d_relighting_comparison_owl}
\end{figure*}
\begin{table}[!t]
    \centering
    \begin{tabular}{lccc}
    \toprule
                                         & \multicolumn{3}{c}{\bfseries {Relighting}}              \\ \cmidrule(lr){2-4}
    \bfseries {Method}                   & {PSNR $\uparrow$} & {SSIM $\uparrow$} & {LPIPS $\downarrow$} \\ \midrule
    \bfseries LightSwitch & \cellcolor{tabsecond}25.43   & \cellcolor{tabfirst}0.84    & \cellcolor{tabsecond}0.297      \\ \midrule
    Mitsuba+NeuS \cite{jakob2022mitsuba3, wang2021neus} & \cellcolor{tabfirst}26.24 & \cellcolor{tabfirst}0.84 & \cellcolor{tabfirst}0.227 \\
    InvRender \cite{zhang2022invrender} & 23.45 & 0.77 & 0.374 \\
    NeRD \cite{boss2021nerd} & 21.71 & 0.65 & 0.540 \\
    NeRFactor \cite{zhang2021nerfactor} & 20.62 & 0.72 & 0.486 \\
    NeROIC \cite{neroic} & 21.59 & \cellcolor{tabthird}0.78 & \cellcolor{tabthird}0.323 \\
    Neural-PIL \cite{boss2021neuralpil} & 19.56 & 0.51 & 0.604 \\
    NVDiffrec \cite{Munkberg_2022_CVPR} & 22.60 & 0.72 & 0.406 \\
    NVDiffrecMC \cite{hasselgren2022nvdiffrecmc} & 20.24 & 0.73 & 0.393 \\
    PhySG \cite{physg2021} & 22.77 & \cellcolor{tabsecond}0.82 & 0.375 \\
    TensoIR \cite{Jin2023TensoIR} & \cellcolor{tabthird}24.15 & 0.77 & 0.378 \\
    MaterialFusion \cite{litman2024materialfusion} & 20.75 & 0.73 & 0.388 \\
    \bottomrule
    \end{tabular}
    \caption{\textbf{Relighting on the Objects With Lighting Dataset.} Our method matches and outperforms a multitude of inverse rendering baselines on novel view relighting across the Objects With Lighting dataset.}
    \label{tab:owl_3d_comparison}
\vspace*{-2mm}
\end{table}
To validate LightSwitch's relighting and denoising scheme, we evaluate with source RGB images of synthetic test objects captured under an unknown illumination and relight them with a target lighting condition not included in the training environment map dataset. We render the objects under the corresponding target illuminations and compare the relit predictions to the groundtruth appearance. We utilize 6 diverse test objects from the BlenderVault dataset excluded from training and render 8 randomly sampled views on a hemisphere with the object at its center. The object appearances are rendered given 3 fixed illuminations for those 8 views, giving a total of 144 test images.

\paragraph{Baselines.} 

We test LightSwitch against other diffusion-based relighting methods~\cite{jin2024neural_gaffer, zeng2024dilightnet} trained to predict relighting for object data from a single image and report the relighting comparison after rescaling. To highlight the relighting consistency across multiple views, in addition to the typical image level rescaling (ILR) metric that searches for optimal rescaling for each image, we also report a stricter scene level rescaling (SLR) metric that computes a single scale across all views in the scene, penalizing inconsistent predictions across views.

\paragraph{Results.} 

We report quantitative results in Tab.~\ref{tab:2d_relighting_comparison}, comparing LightSwitch with previous image relighting methods. Overall, the gains over baselines highlight the benefits of our design choices of leveraging multi-view attention and material priors for relighting objects with complex and diverse materials. In particular, our method achieves the most consistency when using groundtruth materials, indicating its ability to exploit material information for a more consistent and accurate relighting. Fig.~\ref{fig:2d_relighting_comparison} shows a qualitative comparison of our model's relightings against baselines.

\subsection{Relighting for 3D}

We evaluate the novel view relighting quality and efficiency of LightSwitch against prior inverse rendering methods on a number of publicly available diverse synthetic and real object datasets. For synthetic objects, we directly render and compare the groundtruth relightings under novel environment maps, while real objects were captured in novel environments for which the illuminations are estimated.

\paragraph{Datasets.} 

We use the NeRF synthetic dataset (5 objects) \cite{mildenhall2020nerf} and real object dataset Objects with Lighting (8 objects) \cite{Ummenhofer2024OWL}. NeRF synthetic objects are relit by four high resolution environment maps, and the relighting comparison is computed on a test set of eight unseen poses per environment map. For Objects with Lighting, objects are relit with two novel environment maps for six test views, three per environment map. We show our method's relighting quantitatively and qualitatively on both of these datasets to highlight its strong performance on synthetic and real data. 
\begin{figure}
    \centering
    \includegraphics[trim={4cm 2.06cm 2.7cm 3.32cm},clip,width=0.9\linewidth]{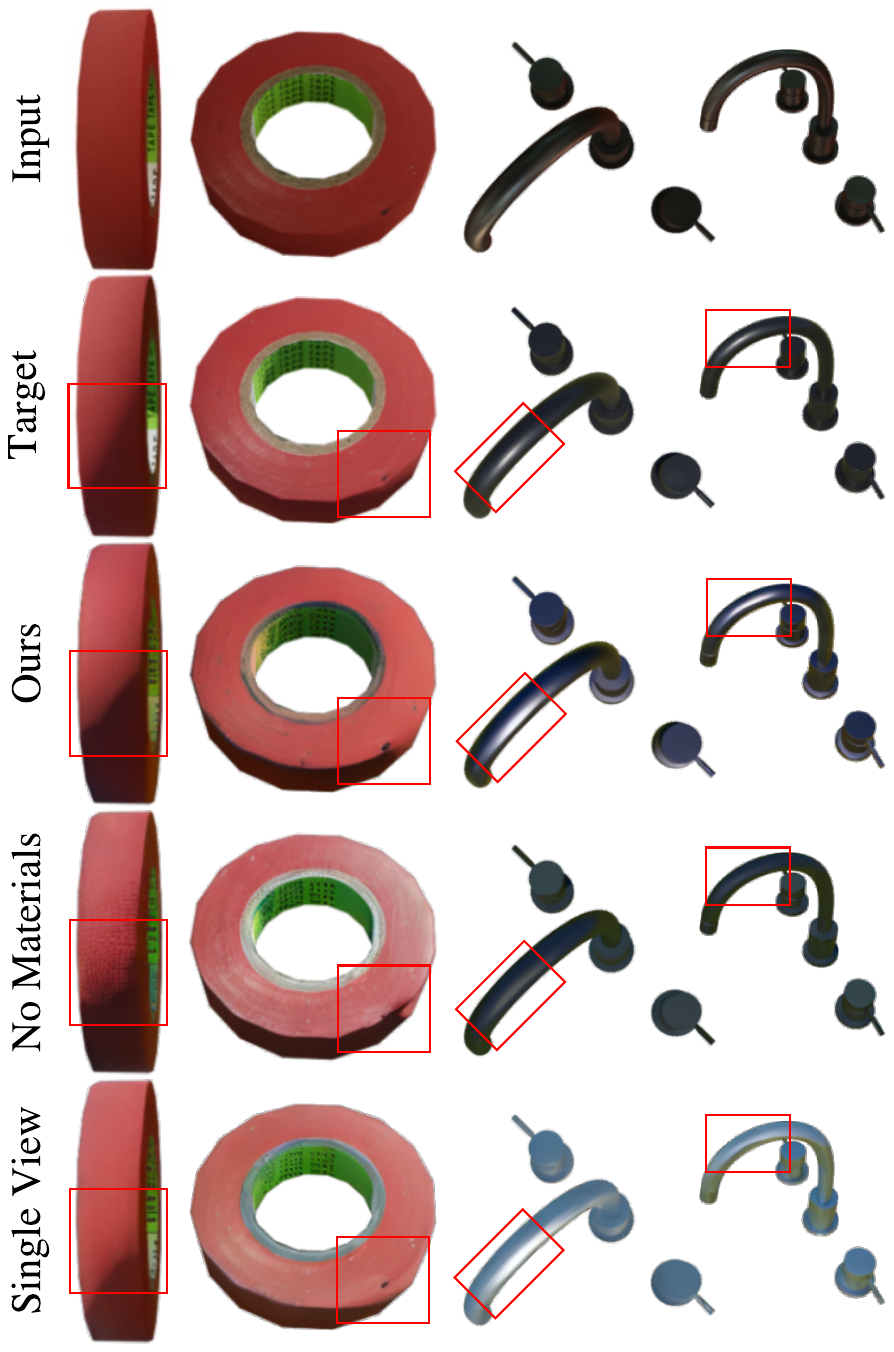}

    \caption{\textbf{2D Relighting Ablation Comparison.} 
    }
\vspace*{-5mm}
    \label{fig:2d_ablations}
\end{figure}

\paragraph{Results.} 
\begin{figure*}
    \centering
    \includegraphics[trim={0.15cm 10.7cm 0.15cm 10.3cm},clip,width=\textwidth]{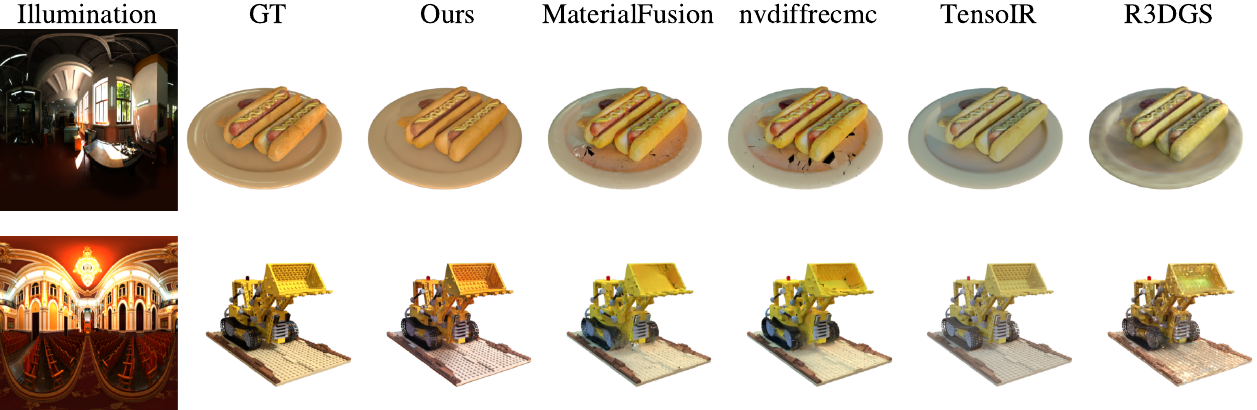}
    \caption{\textbf{3D Relighting Comparison on NeRF-Synthetic.} While other methods exhibit issues in the relit appearance such as baked-in albedo, reconstruction artifacts, and incorrect geometry, our method successfully relights with high fidelity.
    }
    \label{fig:3d_relighting_comparison_nerf}
\end{figure*}
\begin{table*}[!t]
    \centering
\addtolength{\tabcolsep}{-0.45em}
    {\begin{tabular}{lcc cc cc cc cc | cc}\toprule
    & \multicolumn{2}{c}{\bfseries Chair} & \multicolumn{2}{c}{\bfseries Hotdog} & \multicolumn{2}{c}{\bfseries Lego} & \multicolumn{2}{c}{\bfseries Materials} & \multicolumn{2}{c}{\bfseries Mic} & \multicolumn{1}{c}{\bfseries Runtime} \\ 
    \cmidrule(lr){2-3} \cmidrule(lr){4-5} \cmidrule(lr){6-7} \cmidrule(lr){8-9} \cmidrule(lr){10-11} \cmidrule(lr){12-12}
    \bfseries {Method} & PSNR & LPIPS & PSNR & LPIPS & PSNR & LPIPS & PSNR & LPIPS & PSNR & LPIPS & Minutes\\ \midrule
    \bfseries LightSwitch & \cellcolor{tabfirst}26.65 & \cellcolor{tabfirst}0.062 & \cellcolor{tabfirst}25.75 & \cellcolor{tabfirst}0.091 & \cellcolor{tabfirst}23.60 & \cellcolor{tabsecond}0.081 & \cellcolor{tabthird}22.08 & \cellcolor{tabfirst}0.080 & \cellcolor{tabthird}30.24 & \cellcolor{tabsecond}0.025 & \cellcolor{tabfirst}\raisebox{0.5ex}{\texttildelow}2 \\ \midrule
    MaterialFusion \cite{litman2024materialfusion} & \cellcolor{tabsecond}26.58 & \cellcolor{tabsecond}0.063 & \cellcolor{tabsecond}25.31 & \cellcolor{tabsecond}0.123 & \cellcolor{tabthird}23.26 & 0.119 & \cellcolor{tabsecond}25.29 & \cellcolor{tabthird}0.084 & \cellcolor{tabsecond}30.94 & 0.036 & \raisebox{0.5ex}{\texttildelow}240 \\
    NVDiffrecMC \cite{hasselgren2022nvdiffrecmc} & \cellcolor{tabthird}26.44 & \cellcolor{tabthird}0.064 & \cellcolor{tabthird}24.87 & \cellcolor{tabthird}0.133 & \cellcolor{tabsecond}23.36 & \cellcolor{tabthird}0.115 & \cellcolor{tabfirst}25.37 & \cellcolor{tabsecond}0.081 & 30.15 & 0.041 & \cellcolor{tabthird}\raisebox{0.5ex}{\texttildelow}120 \\
    TensoIR \cite{Jin2023TensoIR} & 25.29 & 0.070 & 21.16 & 0.174 & 21.86 & \cellcolor{tabfirst}0.080 & 22.02 & 0.104 & \cellcolor{tabfirst}31.21 & \cellcolor{tabfirst}0.022 & \raisebox{0.5ex}{\texttildelow}480 \\
    R3DGS \cite{R3DG2023} & 23.50 & 0.072 & 21.02 & 0.168 & 20.86 & 0.106 & 20.56 & 0.095 & 29.47 & \cellcolor{tabthird}0.029 & \cellcolor{tabsecond}\raisebox{0.5ex}{\texttildelow}15 \\
    \bottomrule
    \end{tabular}
    }
    
    \caption{
    \textbf{Relighting on the NeRF-Synthetic Dataset.} Our method matches or outperforms the baselines on novel view relighting across all views per NeRF-Synthetic object at a much lower runtime.
    }
    \label{tab:nerf_3d_comparison_quant}
\end{table*}
\begin{table}[!t]
    \centering
    \begin{tabular}{lccc}\toprule
    \bfseries {Ablations} & PSNR $\uparrow$ & SSIM $\uparrow$ & LPIPS $\downarrow$ \\ \midrule
        Ours & \cellcolor{tabfirst}26.01 & \cellcolor{tabfirst}0.888 & \cellcolor{tabfirst}0.216\\
        No Materials & \cellcolor{tabsecond}25.27 & \cellcolor{tabsecond}0.879 & \cellcolor{tabsecond}0.219\\
        Single View & \cellcolor{tabthird}24.59 & \cellcolor{tabthird}0.865 & \cellcolor{tabthird}0.228\\
        \bottomrule
    \end{tabular}
    \caption{\textbf{Effects of Ablating Multi-view or Materials.} Ablating different information from the relighting diffusion framework harms relighting accuracy. 
    }
    \label{tab:ablations}
\end{table}

We benchmark LightSwitch against a suite of inverse rendering baselines on both datasets. As shown in Tab.~\ref{tab:owl_3d_comparison} and Tab.~\ref{tab:nerf_3d_comparison_quant}, our trained model matches or outperforms state-of-the-art methods in relighting across multiple synthetic and real objects. Our model can successfully relight objects exhibiting complex appearance properties. By distributing denoising across compute with our denoising scheme, we relight in much less time than the next best performing baseline, which takes orders of magnitude longer than ours to do relighting. Fig.~\ref{fig:3d_relighting_comparison_nerf} and Fig.~\ref{fig:3d_relighting_comparison_owl} show a qualitative comparison of our model against previous approaches on both datasets. We compare the runtime for our method using 8 RTX A6000 and the other methods that can only utilize 1 RTX A6000 and find that our method is proportionally faster while outperforming or matching baselines. Using 1 GPU increases runtime to 14 minutes without a loss in quality, which is comparable to R3DGS \cite{R3DG2023} in runtime but produces much more accurate relighting. The strong performance in both 2D and 3D relighting exhibited by our method showcases its utilization of multi-view appearance and material cues for effective relighting. 

\subsection{Ablation Studies}

We finetune additional diffusion models that ablate multi-view or material information incorporation into the architecture and evaluate on the BlenderVault 2D test dataset. The quantitative and qualitative comparisons are shown in Tab.~\ref{tab:ablations} and Fig.~\ref{fig:2d_ablations}. Not giving material information during training leads to a considerable drop in quality, as the model struggles with relighting more complex appearances such as specularities, and begins incorporating details from the input views to the predicted relightings. Training with materials but not with multi-view leads to an even bigger drop, as the model struggles to produce consistent relightings.
\section{Conclusion}
\label{sec:conclusion}

In this paper, we introduced LightSwitch, a generative relighting framework capable of leveraging inferred material cues for accurate and consistent multi-view relighting. While this improved over prior works in 2D and 3D relighting, we believe there are some remaining limitations. First, as shown in the appendix, the reliance on the (fixed) latent space of a pre-trained diffusion limits the ability to encode/decode sharp fine details \eg reflections. Moreover, while our architecture encourages multi-view consistency and material-aware inference, the predictions are not guaranteed to be physically plausible. We believe that exploring alternate architectures and mechanisms to more closely connect learned relighting with physics-based rendering are promising avenues for future exploration.

\section{Acknowledgments}
\label{sec:acknowledgments}
This work was supported in part by the NSF GFRP (Grant No. DGE2140739) and NSF Award IIS-2345610. This work used Bridges-2 at Pittsburgh Supercomputing Center through allocation CIS240022 from the ACCESS program, which is supported by National Science Foundation grants \#2138259, \#2138286, \#2138307, \#2137603, and \#2138296.

{
    \small
    \bibliographystyle{ieeenat_fullname}
    \bibliography{main}
}

\clearpage
\setcounter{page}{1}
\maketitlesupplementary

\begin{table}[!t]
    \centering
    \setlength{\tabcolsep}{1pt}
        \centering
        \resizebox{\linewidth}{!}{\begin{tabular}{ccccc}
            \toprule  & \multicolumn{4}{c}{\bfseries {Relighting}} \\ \cmidrule(lr){2-5}
            \textbf{Method} & PSNR-L $\uparrow$ & PSNR-H $\uparrow$ & SSIM $\uparrow$ & LPIPS $\downarrow$ \\ \midrule
            \bfseries LightSwitch & \cellcolor{tabthird}32.02 & \cellcolor{tabthird}25.03   & \cellcolor{tabsecond}0.976    & \cellcolor{tabsecond}0.027      \\ \midrule
            Neural-PBIR  & \cellcolor{tabfirst}33.26 & \cellcolor{tabfirst}26.01 & \cellcolor{tabfirst}0.979 & \cellcolor{tabfirst}0.023 \\
            IllumiNeRF  & \cellcolor{tabsecond}32.74 & \cellcolor{tabsecond}25.56 & \cellcolor{tabsecond}0.976 & \cellcolor{tabsecond}0.027 \\
            NVDiffrecMC  & 31.60 & 24.43 & \cellcolor{tabthird}0.972 & 0.036 \\
            RelitLRM  & 31.52 & 24.67 & 0.969 & \cellcolor{tabthird}0.032 \\
            InvRender & 30.83 & 23.76 & 0.970 & 0.046 \\
            NeRFactor  & 30.38 & 23.54 & 0.969 & 0.048 \\
            NVDiffrec  & 29.72 & 22.91 & 0.963 & 0.039 \\
            \bottomrule
        \end{tabular}}
        \caption{3D Relighting Comparison on Stanford-ORB.}
        \label{tab:stanford_orb_relighting}
\end{table}

\section{Additional Visualizations}
\label{sec:additional_vis}
\begin{figure*}
    \centering
    \includegraphics[width=\textwidth]{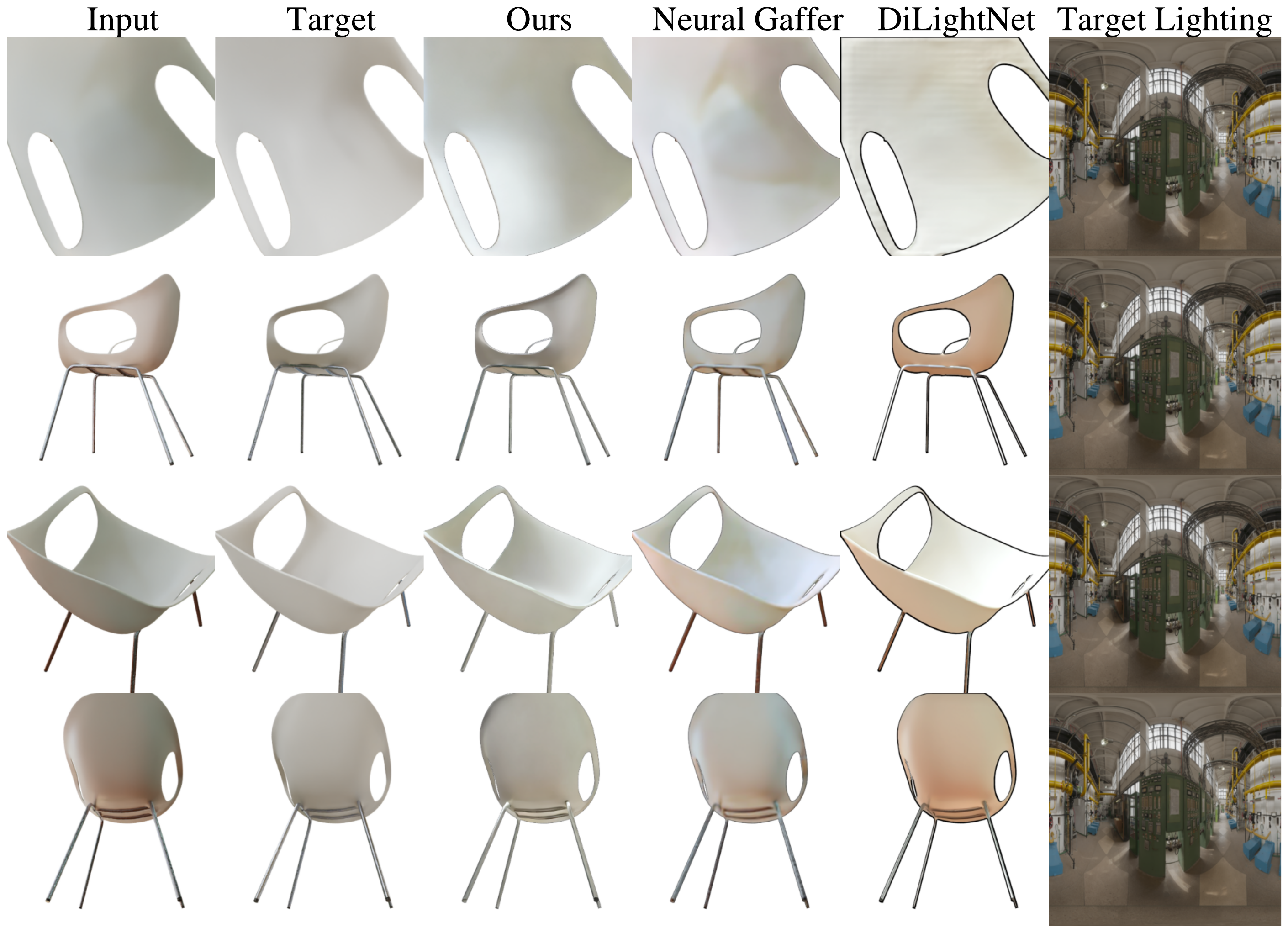}
    \caption{\textbf{Additional Visualizations of LightSwitch Relighting on Synthetic Objects.}
    }
    \label{fig:aerodynamics_workshop-boiler_room_11665_relighting_comparison}
\end{figure*}

\begin{figure*}
    \centering
    \includegraphics[width=\textwidth]{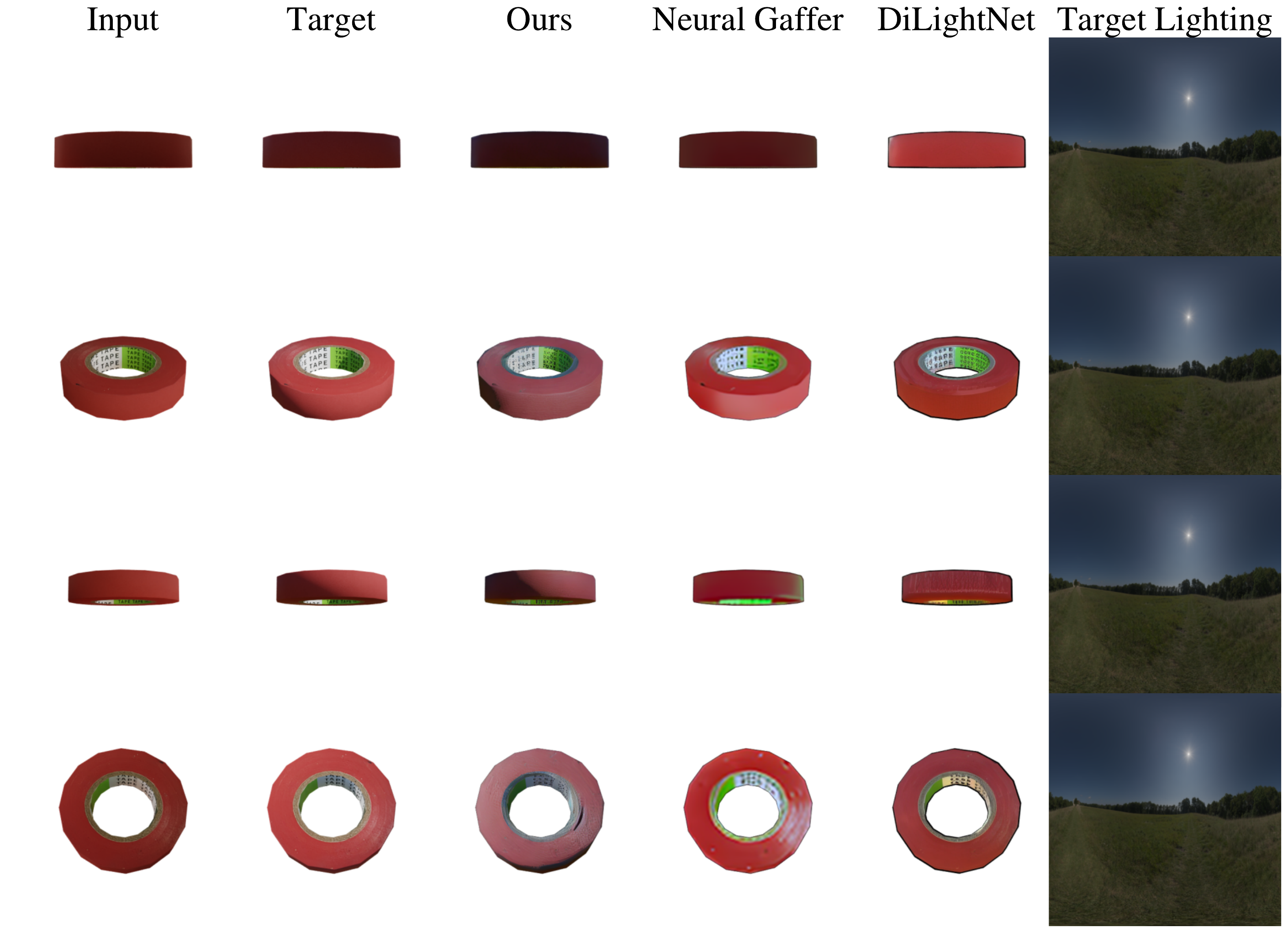}
    \caption{\textbf{Additional Visualizations of LightSwitch Relighting on Synthetic Objects.}
    }
    \label{fig:aerodynamics_workshop-dreifaltigkeitsberg_11664_relighting_comparison}
\end{figure*}

\begin{figure*}
    \centering
    \includegraphics[width=\textwidth]{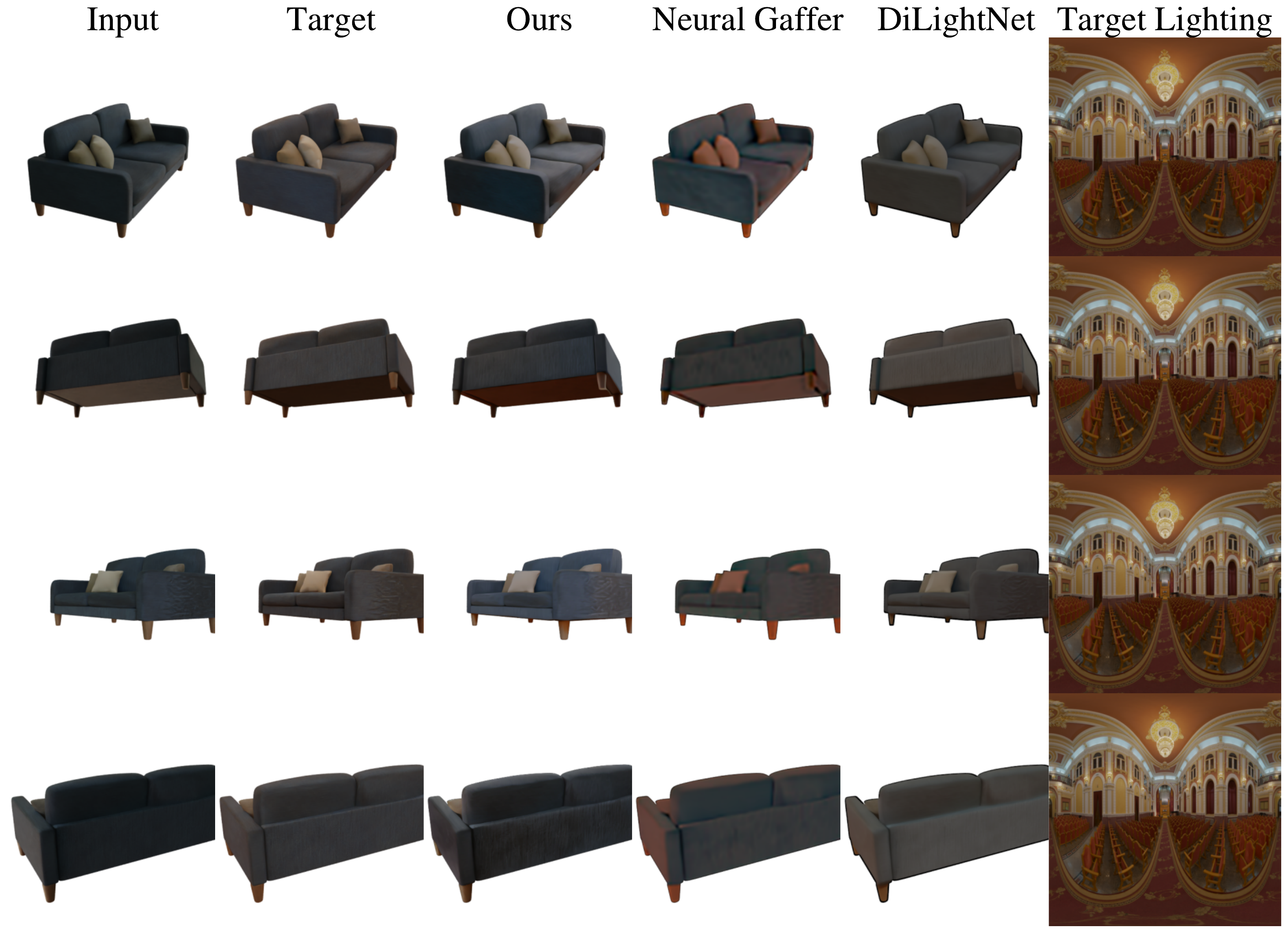}
    \caption{\textbf{Additional Visualizations of LightSwitch Relighting on Synthetic Objects.}
    }
    \label{fig:aerodynamics_workshop-music_hall_11662_relighting_comparison}
\end{figure*}

\begin{figure*}
    \centering
    \includegraphics[width=0.96\textwidth]{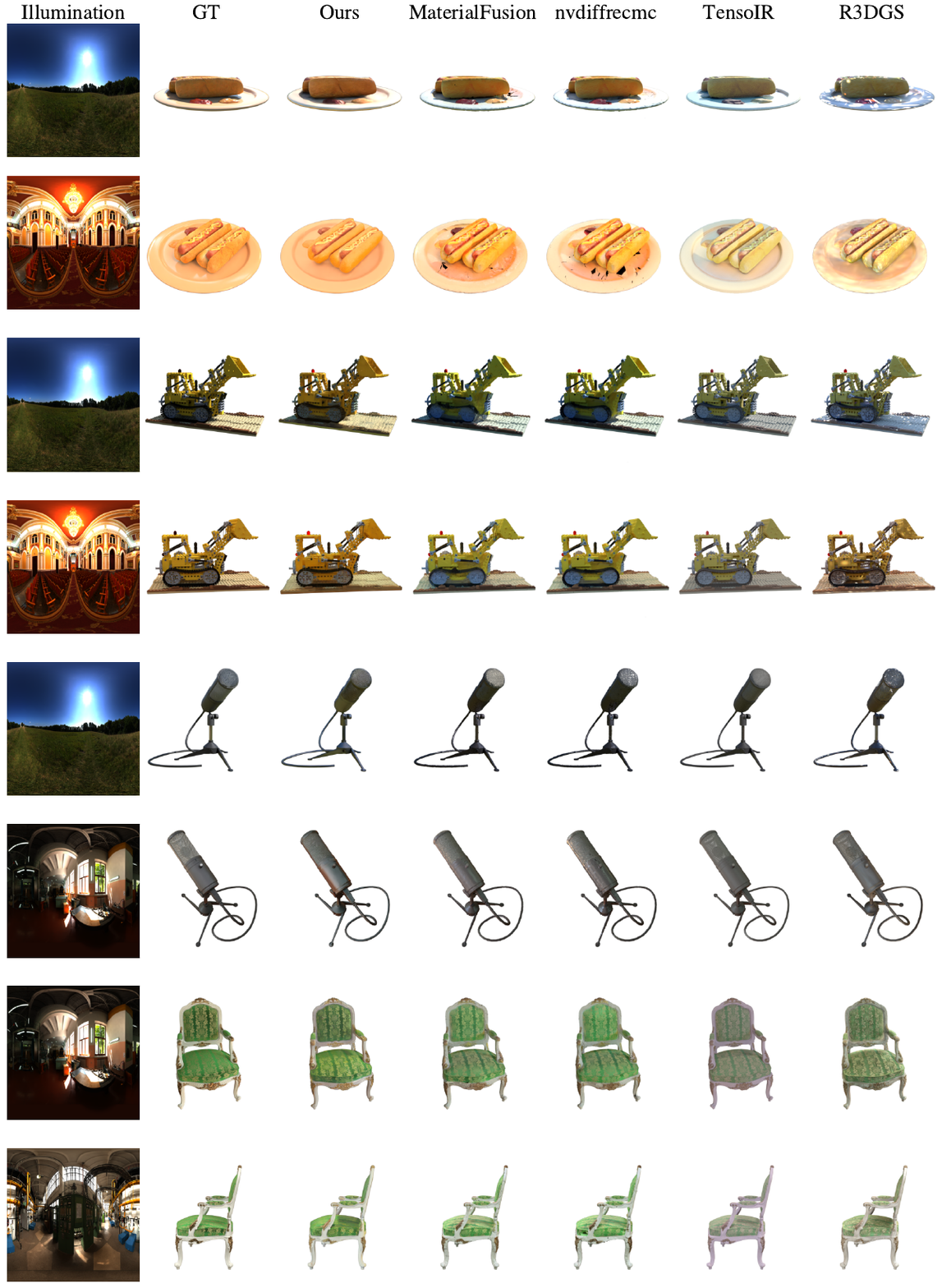}
    \caption{\textbf{Additional Visualizations of LightSwitch 3D Relightings on NeRF-Synthetic.}
    }
    \label{fig:3d_nerf_synthetic_sup}
\end{figure*}

We show additional visualizations of LightSwitch's 2D and 3D relighting on BlenderVault 2D data as well as NeRF-Synthetic in Figs.~\ref{fig:aerodynamics_workshop-boiler_room_11665_relighting_comparison}-\ref{fig:3d_nerf_synthetic_sup}.

\section{Additional Details}
\label{sec:additional_details}

\begin{table*}[!t]
    \centering
\addtolength{\tabcolsep}{-0.45em}
    {\begin{tabular}{lcc cc cc cc cc}\toprule
    & \multicolumn{2}{c}{\bfseries Mic} & \multicolumn{2}{c}{\bfseries Hotdog} & \multicolumn{2}{c}{\bfseries Chair} & \multicolumn{2}{c}{\bfseries Lego} & \multicolumn{2}{c}{\bfseries Materials}  \\ \cmidrule(lr){2-3} \cmidrule(lr){4-5} \cmidrule(lr){6-7} \cmidrule(lr){8-9} \cmidrule(lr){10-11}
    \bfseries {Method} & PSNR & LPIPS & PSNR & LPIPS & PSNR & LPIPS & PSNR & LPIPS & PSNR & LPIPS\\ \midrule
\bfseries LightSwitch & 30.24 & 0.025 & 25.91 & 0.090 & 26.65 & 0.062 & 23.60 & 0.081 & 22.08 & 0.080 \\ \midrule
MaterialFusion \cite{litman2024materialfusion}& 30.46 & 0.045 & 23.09 & 0.153 & 25.40 & 0.084 & 21.87 & 0.145 & 20.47 & 0.158\\
NVDiffrecMC \cite{hasselgren2022nvdiffrecmc}   & 29.81 & 0.052 & 22.88 & 0.159 & 25.39 & 0.083 & 22.04 & 0.141 & 20.50 & 0.157 \\
TensoIR \cite{Jin2023TensoIR}       & 30.92 & 0.024 & 21.12 & 0.179 & 24.82 & 0.082 & 21.37 & 0.100 & 22.01 & 0.107\\
R3DGS \cite{R3DG2023} & 28.87 & 0.033 & 20.89 & 0.179 & 23.08 & 0.084 & 20.38 & 0.129 & 20.48 & 0.101\\
        \bottomrule
    \end{tabular}
    }
    \caption{
    \textbf{Relighting on the NeRF-Synthetic Dataset.} We report the performance of all other baselines when their images are encoded/decoded using the Stable Diffusion encoder and decoder before comparison. The VAE causes significant drops in relighting quality for all objects, especially those with reflections sharp fine reflections such as \texttt{materials}, which explains our method's struggle on the object. 
    }
    \label{tab:nerf_vae_3d_comparison}
\end{table*}

\paragraph{Training.}
LightSwitch was trained in three stages using 8 RTX A6000 GPUs, first by finetuning for single view for 20K iterations on 256$\times$256 data with a batch size of 512. An AdamW 8-bit optimizer was used with a learning rate of $5e-5$. For the second multi-view stage, we train with a batch size of 120, with each batch containing four 256$\times$256 images randomly sampled for the object. This was done for 15K iterations with a learning rate of $2.5e-5$. For the last upsampling stage, a batch size of 28 was used, where each batch contained four 512$\times$512 images. This was done with a learning for 15K iterations with a learning rate of $1e-5$. We repeat the upsampling stage training for StableMaterialMV in order to create higher quality material maps.

\paragraph{Stanford-ORB Relighting Evaluation.}
We report results on Stanford-ORB in Tab.~\ref{tab:stanford_orb_relighting}, showing LightSwitch is competitive with SOTA while performing significantly faster -- relighting a scene in 8 minutes vs. several hours for baselines. Due to lighting changes when moving the capture device, Stanford-ORB has \emph{separate} environment maps for each test view. As our multi-view method relights query views under a common illumination, we disregard the variation and assume the lighting for the first image but this may be suboptimal.

\end{document}